\documentclass[journal]{IEEEtran}

\usepackage{comment}
\usepackage{mathtools}
\usepackage{amsmath}
\usepackage{graphicx}
\usepackage{epstopdf}

\usepackage{hyperref} 
\usepackage{cite}
\usepackage{amsmath}
\usepackage{color}

\usepackage{graphicx}
\usepackage{mathptmx}      
\usepackage{latexsym}
\usepackage[ruled,vlined]{algorithm2e}

\usepackage[misc]{ifsym}

\DeclareMathOperator{\atantwo}{atan2}

\begin{document}

\title{StereoNeuroBayesSLAM: A Neurobiologically Inspired Stereo Visual SLAM System Based on Direct Sparse Method}

\author{Taiping~Zeng, Xiaoli~Li,
and~Bailu~Si~\IEEEmembership{Member,~IEEE}
\thanks{T. Zeng is with Institute of Science and Technology for Brain-Inspired  Intelligence, Fudan University, Shanghai, China and Key Laboratory of Computational Neuroscience and Brain-Inspired Intelligence (Fudan University), Ministry of Education, China (e-mail:zengtaiping.ac@gmail.com).}
\thanks{X. Li is with State Key Laboratory of Cognitive Neuroscience and Learning, Beijing Normal University, Beijing 100875, China (e-mail:xiaoli@bnu.edu.cn).}
\thanks{B. Si is with School of Systems Science, Beijing Normal University, 100875, China (e-mail:bailusi@bnu.edu.cn).}
\thanks{Correspondence should be addressed to Bailu Si (bailusi@bnu.edu.cn).}
}


\maketitle

\begin{abstract}
We propose a neurobiologically inspired visual simultaneous localization and mapping (SLAM) system based on direction sparse method to real-time build cognitive maps of large-scale environments from a moving stereo camera. 
The core SLAM system mainly comprises a Bayesian attractor network, which utilizes neural responses of head direction (HD) cells in the hippocampus and grid cells in the medial entorhinal cortex (MEC) to represent the head direction and the position of the robot in the environment, respectively. Direct sparse method is employed to accurately and robustly estimate velocity information from a stereo camera.
Input rotational and translational velocities are integrated by the HD cell and grid cell networks, respectively.
We demonstrated our neurobiologically inspired stereo visual SLAM system on the KITTI odometry benchmark datasets. Our proposed SLAM system is robust to real-time build a coherent semi-metric topological map from a stereo camera. Qualitative evaluation on cognitive maps shows that our proposed neurobiologically inspired stereo visual SLAM system outperforms our previous brain-inspired algorithms and the neurobiologically inspired monocular visual SLAM system both in terms of tracking accuracy and robustness, which is closer to the traditional state-of-the-art one.
\end{abstract}


%
\IEEEpeerreviewmaketitle

\section{Introduction}
\label{intro}

\IEEEPARstart{M}{any} animals, including humans, can freely navigate the large-scale, dynamic, complex environment over thousands of miles for a long-lasting period. These are the basic abilities of animals including exploration, localization, cognitive mapping, and navigation, called spatial cognition.
Tolman proposes the first explanation that an internal map-like representation, i.e., cognitive map, guides animals to perceive space and travel in an environment, and represents the spatial relationship between salient landmarks and animals, and the spatial relationship among salient landmarks \cite{tolman_cognitive_1948}. 
The discovery of place cells embodies the existence of the cognitive map~\cite{okeefe_hippocampus_1971,okeefe_hippocampal_1978}. 
Place cells in the hippocampus fire selectively only when animals at one or a few locations in the environment. Different place cells have different firing locations, called place fields. The population coding of place cells can represent not only the current location of animals, but also the earlier locations~\cite{moser2015place}. 
Ensembles of HD cells represent the animal's head direction within an allocentric reference frame, like a compass~\cite{taube_head-direction_1990, taube_head_2007}.

The historical discovery of grid cells in the dorsolateral band of the medial entorhinal cortex (dMEC) builds a bridge between neuroscience evidence and computational neural models~\cite{hafting_microstructure_2005}. The grid cells in dMEC periodically fire in multiple locations in the environment. The firing fields of grid cells form a hexagon grid pattern spanning the whole explored environment. It is widely believed that the metric representation of space by grid cells provides spatial information to place cells involved in the formation of cognitive maps~\cite{mcnaughton_path_2006}. Ensembles of grid cells can do accurate path integration of the animal's velocity to help hippocampal formation encoding relative spatial location, without reference to external salient landmarks~\cite{moser_place_2008, yoram_burak_accurate_2009}.
As accurate path integration needs precise velocity inputs, rough velocity estimation can only generate inaccurate metric representation to build distorted cognitive maps.

Many computational neural models of HD cells, place cells, and grid cells are validated by robot navigation system~\cite{burgess_robotic_1997, strosslin_robust_2005, jauffret_grid_2015, zeng_cognitive_2017, zeng_global_2017}. It promotes a deeper understanding of how spatial cognition works in the brain. What is more important is that neurobiological evidence has inspired to develop more pragmatic navigation systems, typically like openRatSLAM~\cite{ball_openratslam:_2013}. In the brain-inspired RatSLAM model, abstract pose cells are proposed to represent the position and head direction of the robot. Pose cell activity is updated by displacing a copy of the activity pump. Local view cells are introduced to perceive vision inputs and a cognitive map is built according to the neural codes of the pose cells. OpenRatSLAM can build a coherent cognitive map in the large-scale environment using a single web camera.
For pure visual inputs, current visual odometry of RatSLAM system is only able to make a rough estimation of translational and rotational velocity from two consecutive images, which mainly depends on loop closures to correct estimation errors. 

However, grid cells perform accurate path integration only based on accurate velocity estimation. If the loop is closed after a long trip for long time intervals, it would lead to severe distortion of cognitive map in the large-scale environment. The distortion cannot be easily eliminated by the optimization of the topological map.  

More recently, an efficient visual SLAM solution based on ORB features, ORB-SLAM, has been developed~\cite{mur-artal_orb-slam:_2015, mur-artal_orb-slam2:_2017}. 
Another benchmark visual odometry systems are Direct Sparse Odometry (DSO) and stereo DSO, which are shown to outperform the state-of-the-art methods for visual SLAM in terms of both tracking accuracy and robustness~\cite{engel_direct_2017, wang_stereo_2017}. Direction sparse method for visual odometry has achieved great performance with similar accuracy, compared with traditional features based methods. 

However, DSO and stereo DSO are not complete SLAM systems yet, but a visual odometry, and lack functions of loop closure detections, map reuse, and relocalization. Moreover, the open-source DSO is monocular visual odometry, whose scale drift is a significant source of error by far. DSO also suffers from fast motion and rolling shutter cameras.

Neurobiologically inspired robot navigation system has long been considered mainly to test neurobiology hypotheses and validate computational neural models.
In this work, jointing our previous work Bayesian attractor network~\cite{zeng_neurobiologically_2017} and stereo DSO~\cite{wang_stereo_2017}, we proposed a neurobiologically inspired stereo visual SLAM system with direct sparse visual odometry method for highly accurate and robust velocity estimation. The key components of the proposed SLAM system are visual odometry, local view cells, Bayesian attractor network, and cognitive map. Visual odometry adopts the direct sparse method to estimate velocity information from a stereo camera. Local view cells are used to determine whether the current view is familiar or not. In the Bayesian attractor network, probabilistic methods are utilized to represent the ring attractor neural responses of HD cells in the hippocampus and the torus attractor neural manifold of grid cells in the MEC corresponding to the head orientation and the regular space locations of the robot in the environment, respectively. Global optimization of cognitive map after loop closures is implemented by non-linear optimization using Ceres Solver~\cite{agarwal2012ceres, zeng_compactmapping_2017}.
Our proposed neurobiologically inspired SLAM system is demonstrated on the KITTI odometry benchmark dataset~\cite{Geiger2012CVPR}, which is able to real-time build a coherent semi-metric topological map. Quantitative evaluation on cognitive maps shows that our proposed neurobiologically inspired stereo visual SLAM system outperforms our previous brain-inspired algorithms~\cite{zeng_neurobiologically_2017} and the neurobiologically inspired monocular visual SLAM system both in terms of tracking accuracy and robustness in the large-scale outdoor environment.
This work makes the following specific contributions:
\begin{itemize}
\item A neurobiologically inspired stereo visual SLAM system is modularly implemented by Robot Operating System (ROS) with visualization.
\item More accurate velocity is provided by direct sparse method to the HD cell and grid cell networks for path integration.
\item Stereo DSO and DSO are expanded to a full SLAM system.
\item The proposed neurobiologically inspired stereo visual SLAM system is demonstrated on the KITTI vision benchmark datasets.
\item Qualitative comparisons to our previous SLAM system improved from RatSLAM and the neurobiologically inspired monocular visual SLAM system with direct sparse method, the neurobiologically inspired stereo visual SLAM system is apparently superior to these methods in terms of both accuracy and robustness.
\end{itemize}

This work extends from our previous works~\cite{zeng_neurobiologically_2017, zeng_compactmapping_2017} as these provide detailed information about Bayesian attractor network and non-linear optimization of topological map. We present direct sparse visual odometry, Bayesian attractor network, non-linear optimization of topological map in Section 2. Neural activities of HD cells and grid cells, cognitive map, firing rate maps, and map evaluations are presented in Section 3. Experimental results are discussed in Section 4. Section 5 gives a brief conclusion.

\section{Method}
\label{Method}

In this study, we build a neurobiologically inspired stereo visual SLAM system with real-time, accurate and robust performance in the large-scale environment, which mimics spatial cognitive ability in the entorhinal-hippocampal circuits of mammalian brains. 
Bayesian attractor network model integrates different sensory modalities under uncertainty.
Vestibular and visual cues are received by the ventral intraparietal (VIP) area and the dorsal medial superior temporal (MSTd) area, respectively. These two inputs are probabilistically integrated on the HD cell layers to represent the animal's head direction. The similar probabilistic mechanism of HD cells network are applied to the grid cell network for the encodings of periodic spacing locations.
The vestibular cue is provided by a proved direct sparse visual odometry for achieving good performance both in terms of accuracy and robustness~\cite{wang_stereo_2017}.
The Local view cells analogous to the visual cortex provide visual cues to the Bayesian attractor network.
Cognitive map acts as the role of the hippocampus in the brain.
Here, we describe the core components of the SLAM system, and how this component can shape into a high-performance vision-only SLAM system for the large-scale environment. 

\subsection{Network Architecture}

\begin{figure*}[!ht]
\centering
\includegraphics[width=6.8in]{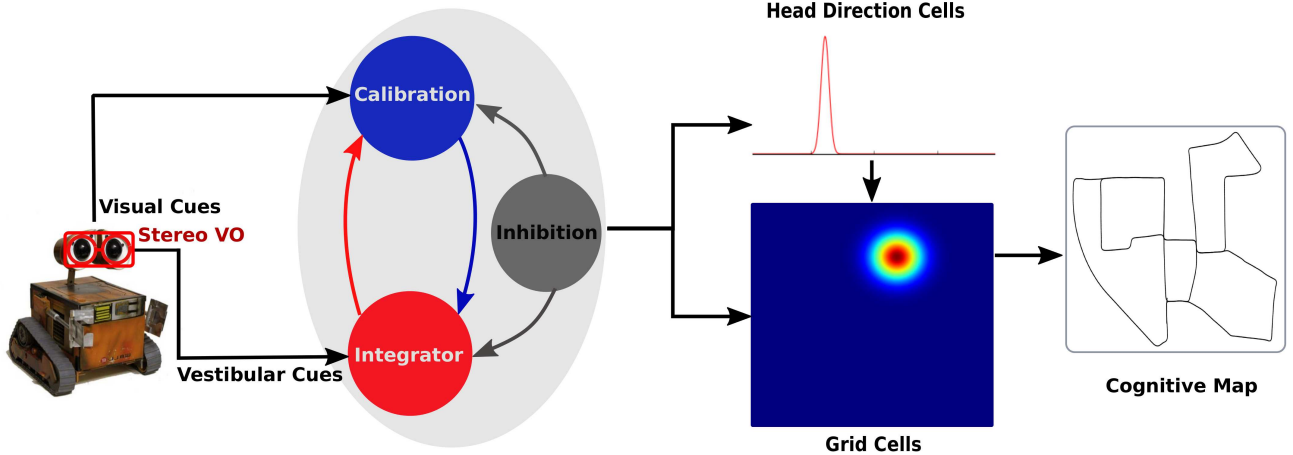}
\caption{Bayesian attractor network architecture of the model and information flow char. Vestibular cues are achieved from a stereo visual odometry represented by integrator cells. Visual cues are presented by calibration cells. Conflict between cues is modeled by mutual inhibition. Global inhibition forms a single stable peak. Integrator cells and calibration cells are together to form the neural presentation of HD cells and grid cells. Head direction is encoded by HD cells. Locations are represented by grid cells with inputs from HD cells, which finally generates a cognitive map.}
\label{fig_arch}
\end{figure*}

The architecture of the prosed model is shown in Fig.~\ref{fig_arch}.
Rotational and translational velocity is estimated by a direct sparse visual method as vestibular cues. The velocity estimation method is not a brain-inspired one, but a highly accurate and robust visual odometry. This visual odometry can make all our SLAM system run in real-time on a standard desktop computer. The direct sparse visual odometry combines static stereo vision with temporal multi-view stereo vision to break the limited accurate depth range caused by the fixed baseline.

In the Bayesian attractor network, a probabilistic based model is proposed to represent the grid firing pattern and HD firing bumps~\cite{zeng_neurobiologically_2017}. The grid cell network model is an extended HD cell network model. Integrator cells, calibration cells, and inhibition cells are included in the model. Integrator cells are presented to integrate vestibular cues. Visual cues are modeled by calibration cells. Cue conflicts are modeled by mutual inhibition. Global inhibition leads to form a stable firing state of the neural network. Bayesian integration eliminates cue uncertainties.

Eventually, with accurate velocity estimation, precise path integration is performed by HD cells network and grid cell network to represent the head direction and position of the robot, and cognitive map is generated depending on a series of experiences when the robot explores in the environment.

\subsection{Bayesian Attractor Network}
The Bayesian attractor neural network model is a model based on Bayesian theory to represent ring attractor neural responses of HD cells by a one-dimensional Gaussian distribution with periodic boundary conditions, and the torus attractor neural manifold of grid cells with a single peak achieving by a two-dimensional Gaussian activity packets with periodic boundary conditions. We provide an overview of the ring attractor model of HD cells, and the torus attractor model of grid cells.

\subsubsection{Head Direction Cell Model}
Rotation of the robot is modeled by the head direction cell network, which has the same angular velocity inputs of the robot in the physical environment.
The neural manifold of the HD cell network is updated by attractor dynamics, vestibular cues integration, and visual cues calibration. The ring attractor manifold limits the HD phase $\theta$ to a range $[0,2\pi)$.

\paragraph{Attractor Dynamics}
The integrator cell and calibration cell are described by the normal distribution
\begin{equation}
f(x) = \frac{1}{{\sigma \sqrt {2\pi } }}e^{{{ - \left( {x - \mu } \right)^2 } / {2\sigma ^2 }}}.
\label{eq:normalDistribution}
\end{equation}
Mutual inhibition between integrator cell and calibration cell and global inhibition are key to achieve the attractor dynamic. The neural manifold is evolved continuously over time, and eventually form a stable neural activity without energy inputs. 
The global inhibition can be defined by 
\begin{equation}
\begin{split}
\displaystyle
\frac{1}{\sigma_{\text{cc}}^{t}{}^2} &= \frac{1}{\sigma_{\text{inte}}^{t-1}{}^2} + \frac{1}{\sigma_{\text{cali}}^{t-1}{}^2}\\
\frac{1}{\sigma_{\text{inte}}^{t}{}^2} &= E\,\,\frac{\frac{1}{\displaystyle \sigma_{\text{inte}}^{t-1}{}^2}}{\frac{1}{\displaystyle \sigma_{\text{cc}}^{t}{}^2} }\\
\frac{1}{\sigma_{\text{cali}}^{t}{}^2} &= E\,\,\frac{\frac{1}{\displaystyle \sigma_{\text{cali}}^{t-1}{}^2}}{\frac{1}{\displaystyle \sigma_{\text{cc}}^{t}{}^2} },
\end{split}
\end{equation}
where $\displaystyle \frac{1}{\sigma_{\text{inte}}^{t-1}{}^2}$ and $ \displaystyle \frac{1}{\sigma_{\text{cali}}^{t-1}{}^2}$ are the previous weight of integrator cell and calibration cell, respectively. $\displaystyle \frac{1}{\sigma_{\text{cc}}^{t}{}^2}$ is the sum of previous weight. $\displaystyle \frac{1}{\sigma_{\text{inte}}^{t}{}^2}$ and $ \displaystyle \frac{1}{\sigma_{\text{cali}}^{t}{}^2}$ are the current weight. $E$ is a constant, which is the total neural activity energy. The mutual inhibition can be described by 
\begin{equation}
\begin{split}
\displaystyle
\frac{1}{\sigma_{\text{inte}}^{t}{}^2} &= \frac{1}{\sigma_{\text{inte}}^{t-1}{}^2} - 
\Delta_{\text{inte}}\,\,\frac{1}{\sigma_{\text{cali}}^{t-1}{}^2}\\
\frac{1}{\sigma_{\text{cali}}^{t}{}^2} &= \frac{1}{\sigma_{\text{cali}}^{t-1}{}^2} - \Delta_{\text{cali}}\,\,\frac{1}{\sigma_{\text{inte}}^{t-1}{}^2},
\end{split}
\end{equation}
where $\Delta_{\text{inte}}$ and $\Delta_{\text{cali}}$ are the mutual inhibition intensity to each other.

\paragraph{Vestibular Cues Integration}
Vestibular cues are integrated by simply shifting the mean of the normal distribution without bump deformation during the process of path integration.
Path integration can be implemented by 
\begin{equation}
\begin{split}
\mu_{\text{inte}}^{t} &=\mod(\mu_{\text{inte}}^{t-1} + \nu^{t} \Delta t, 2\pi)\\
\mu_{\text{cali}}^{t} &=\mod(\mu_{\text{cali}}^{t-1} + \nu^{t} \Delta t, 2\pi),
\end{split}
\end{equation}
where $\mu_{\text{inte}}^{t}$ and $\mu_{\text{cali}}^{t}$ are the mean of integrator cell and calibration cell, $\nu^{t}$ is current velocity, $\Delta t$ is the time interval between $t$ and $t-1$. 

\paragraph{Visual Cues Calibration}
Familiar visual cues can calibrate the neural activity of HD cells. When the current view is novel, a new view template is extracted and associated with a new local view cell with a strong one-shot learned link to the current HD pattern. When the robot meets a familiar scene, the local view cell can be reactivated. Energy can be injected into HD cells network through learned link. The energy injection can be written as 
\begin{equation}
\begin{split}
\displaystyle
\frac{1}{\sigma_{\text{cali}}^{t}{}^2} &= \frac{1}{\sigma_{\text{cali}}^{t-1}{}^2} + \frac{1}{\sigma_{\text{inject}}^{t}{}^2}\\
\mu_{\text{cali}}^{t} &= \mod\left(\left( \frac{\mu_{\text{cali}}^{t-1}}{\sigma_{\text{cali}}^{t-1}{}^2} + \frac{\mu_{\text{inject}}^{t}}{\sigma_{\text{inject}}^{t}{}^2} \right) \,\, \sigma_{\text{cali}}^{t}{}^2, 2\pi\right),
\end{split}
\end{equation}
where $\displaystyle \frac{1}{\sigma_{\text{inject}}^{t}{}^2}$ is the intensity of the current injected energy, $\displaystyle \mu_{\text{inject}}^{t}$ is the injected location on the one dimensional neural manifold of HD cells. 

\paragraph{Phase Estimation}
The current HD phase can be estimated from the integrator cell and the calibration cell, whose probabilistic distribution can be described by
 \begin{equation}
\begin{split}
\displaystyle
\frac{1}{\sigma_{\text{cc}}^{t}{}^2} &= \frac{1}{\sigma_{\text{inte}}^{t}{}^2} + \frac{1}{\sigma_{\text{cali}}^{t}{}^2}\\
\mu_{\text{cc}}^{t} &= \mod\left(\left( \frac{\mu_{\text{inte}}^{t}{}}{\sigma_{\text{inte}}^{t}{}^2} + \frac{\mu_{\text{cali}}}{\sigma_{\text{cali}}^{t}{}^2} \right) \,\, \sigma_{\text{cc}}^{t}{}^2, 2\pi\right),
\end{split}
\end{equation}
where $\displaystyle \frac{1}{\sigma_{\text{cc}}^{t}{}^2}$ and $\displaystyle \mu_{\text{cc}}^{t}$ are the estimated wight and HD phase, respectively. If $\displaystyle |\mu_{\text{cc}}^{t} - \mu_{\text{cali}}^{t}| < \text{Threshold}$, the decision that the current view is familiar is made, and HD phase $\displaystyle \mu_{\text{cc}}^{t}$ is assigned to $\displaystyle \mu_{\text{inte}}^{t}$. If it not meets the condition, it would go to the next cycle.

\subsubsection{Grid Cell Model}
The torus manifold of grid cells can be expanded from the ring manifold of HD cells. The same mechanism of the HD cell network is adopted for location representation.
The integrator cell and calibration cell in the torus manifold can be written as
\begin{equation}
\displaystyle
f(x,y) = \frac{1}{{2\pi \sigma_x \sigma_y}}e^{ - [ \left( {x - \mu_x } \right)^2 / {2\sigma_x ^2 } + \left( {y - \mu_y } \right)^2 / {2\sigma_y ^2 } ] }.\label{eq:2DnormalDistribution}
\end{equation}
Same mechanisms in the HD cell network can be used to grid cell representation integrating linear velocity and sensory cues, which also include attractor dynamics, vestibular cues integration, visual cues calibration, phase estimation.
We will not go further into the issue here.

\subsection{Direct Sparse Stereo Visual Odometry}

\begin{figure}[!ht]
\centering
\includegraphics[width=3.4in]{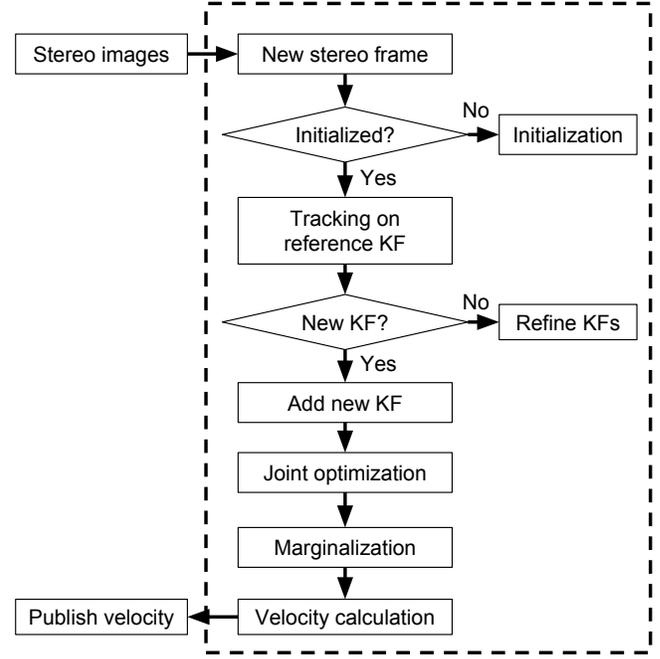}
\caption{The diagram of the direct sparse stereo visual odometry.}
\label{fig_vo_diag}
\end{figure}

The inaccurate estimation of velocity may lead to a severely distorted cognitive map. Previous works are only able to make rough velocity estimation with the scanline intensity profiles~\cite{ball_openratslam:_2013,zeng_cognitive_2017}. A proved visual odometry with more accuracy and robustness~\cite{engel_direct_2017, wang_stereo_2017} is adopted to our SLAM system. We only provide an overview of the visual odometry.

Once the stereo images are fed into the visual odometry, current velocity is estimated and publishes to other nodes by ROS message types.
The diagram of the direct sparse stereo visual odometry is shown in Fig~\ref{fig_vo_diag}.
First, depth estimation from static stereo matching is used to initialize the system.
Second, new stereo frames are tracked with respect to their reference keyframe (KF).
Third, the system determines whether a new keyframe is added to the current active window. If not, a non-keyframe is created to refine the inverse depth of selected points, otherwise a new keyframe is created and added to the current active window. 
Fourth, a joint optimization is performed for all keyframes in the current active window, including keyframes' poses, affine brightness parameters, the
depths of all the observed selected points, and camera intrinsics. Active points not observed by the two latest keyframes and hosted in the old keyframe, as well as the old keyframe are marginalized out to prevent the growth of the size of the active window.
Finally, the current velocity of the new keyframe is estimated after the joint optimization.

In this section, the same denotation in~\cite{engel_direct_2017} are used. Light, bold lower-case letters and bold upper-case letters are used to denote scalars ($u$), vectors (\textbf{t}) and matrices (\textbf{R}) respectively. Light upper-case letters are used to represent functions ($I$).
Camera calibration matrices are denoted by \textbf{K}. Camera poses are represented by matrices of the special Euclidean group $\mathbf{T}_i \in SE(3)$, which transform a 3D coordinate from the camera coordinate system to the world coordinate system. $\mathrm{\Pi}_{\mathbf{K}}$ and $\mathrm{\Pi}_{\mathbf{K}}^{\mathbf{-1}}$ are used to denote camera projection and back-projection functions. 

\subsubsection{Direct Image Alignment Formulation}
Suppose that there is a point set $\mathcal{P}$ in the reference frame $I_{i}$ observed in another frame $I_{j}$, the energy function of direct image alignment can be described as
\begin{equation}
\displaystyle
E_{ij} = \sum_{\mathbf{p} \in \mathcal{P}_{i}} \sum_{\mathbf{\tilde{p}} \in \mathcal{N}_{\mathbf{p}}} \omega_{\mathbf{\tilde{p}}} \left\lVert I_{j}[\mathbf{\tilde{p}^{\prime}}] - b_{j} - \frac{e^{a_{j}}}{e^{a_{i}}}\left(I_{i}[\mathbf{\tilde{p}}] - b_{i} \right) \right\rVert_{\gamma},
\label{eq:direct_align_energyfunc}
\end{equation}
where $\mathbf{p}$ is an image coordinate of a 3D point, $\mathcal{N}_{\mathbf{p}}$ is the 8-point pattern of $\mathbf{p}$, $||\cdot||_{\gamma}$ is a Huber norm, and $a_i, b_i, a_j, b_j$ model an affine brightness change for frame $i$ and $j$. The pattern point $\mathbf{\tilde{p}}$ is projected into $\mathbf{\tilde{p}^{\prime}}$ in $I_{j}$ calculated by
\begin{equation}
\displaystyle
\mathbf{\tilde{p}^{\prime}} = \mathrm{\Pi}_{\mathbf{K}} \left(\mathbf{T}_{ji} \mathrm{\Pi}_{\mathbf{K}}^{\mathbf{-1}}(\mathbf{\tilde{p}},d_{\mathbf{\tilde{p}}}) \right),
\end{equation}
where $d_{\mathbf{\tilde{p}}}$ is the inverse depth of $\mathbf{\tilde{p}}$, $\mathbf{T}_{ji}$ transforms a point from frame $i$ to frame $j$.
$\omega_{\mathbf{\tilde{p}}}$ is a down-weights high image gradients
\begin{equation}
\displaystyle
\omega_{\mathbf{\tilde{p}}} = \frac{c^2}{c^2 + \left\lVert \nabla I_{i}(\mathbf{\tilde{p}}) \right\rVert_{2}^{2}},
\end{equation}
with some constant $c$.

\subsubsection{Tracking}
To initialize the visual odometry system, for the first frame, static stereo matching is used to estimate a semidense depth map. The inverse depth value of points is needed to track the second frame by equation (\ref{eq:direct_align_energyfunc}).

For tracking a new stereo frame each time, all the points inside the active window are projected into the new stereo frame. The optimization is performed by minimizing the energy function of direct image alignment (\ref{eq:direct_align_energyfunc}) with Gauss-Newton. The pose of the new stereo frame is estimated by fixing the depth values.

\subsubsection{Frame Management}
After successfully tracking a new stereo frame, whether making a new keyframe is determined by scene or illumination changes. Scene changing between the new stereo frame and the last keyframe in the active window is evaluated by the mean squared optical flow, and quantized by the relative brightness.

If a new keyframe is required, a sparse set of points, called candidate points, is selected from the current image. In order to select points with sufficient gradient across the images, the image is divided into small blocks, and the size of small blocks is proportional to the image size. An adaptive threshold is calculated in each small block. If the gradient of a point surpasses the threshold of the block, the point would be selected.

If the scene changing is not sufficient, the non-keyframes is used to refine the inverse depth of candidate points. Static stereo matching is used to obtain a better depth initialization for increasing the tracking accuracy.

To prevent the growth of the size of the active window, old keyframes would be marginalized. The old active points hosted in the old keyframes and not observed by the two latest keyframes would be removed. After old keyframes and old points are removed, candidate points become active points hosted in the new keyframe and observed in another keyframe, which leads to create new photometric energy function items for the next step joint optimization.

\subsubsection{Joint Optimization}

In the joint optimization, combination of static stereo and temporal multi-view stereo puts all energy items together. The final energy function can be described as
\begin{equation}
\displaystyle
E = \sum_{i \in \mathcal{F}} \sum_{\mathbf{p} \in \mathcal{P}_{i}} \left( \sum_{j \in obs^t(\mathbf{p})} E_{ij}^{\mathbf{p}} + \lambda E_{is}^{\mathbf{p}} \right),
\label{eq:joint_opt_energyfunc}
\end{equation}
where $\mathcal{F}$ is the set of keyframes in the current window, $\mathcal{P}_i$ is the set of points in the frame $I_i$, $obs^t(\mathbf{p})$ are the observations of $\mathbf{p}$ from temporal multi-view stereo. $\lambda$ is a coupling factor between temporal multi-view stereo and static stereo.  
The energy function of temporal multi-view stereo $E_{ij}^{\mathbf{p}}$ can be described as
\begin{equation}
\displaystyle
E_{ij}^{\mathbf{p}} = \omega_{\mathbf{p}} \left\lVert I_{j}[\mathbf{p^{\prime}}(\mathbf{T}_i,\mathbf{T}_j,d,\mathbf{c})] - b_{j} - \frac{e^{a_{j}}}{e^{a_{i}}}\left(I_{i}[\mathbf{p}] - b_{i} \right) \right\rVert_{\gamma},
\end{equation}
and the energy function of static stereo $E_{is}^{\mathbf{p}}$ can be described as
\begin{equation}
\displaystyle
E_{is}^{\mathbf{p}} = \omega_{\mathbf{p}} \left\lVert I_{i}^{R}[\mathbf{p^{\prime}} ( \mathbf{T}_{ji}, d, \mathbf{c})] - b_{i}^{R} - \frac{e^{a_{j}^{R}}}{e^{a_{i}^{L}}}\left(I_{i}[\mathbf{p}] - b_{i}^{L} \right) \right\rVert_{\gamma},
\end{equation}
where $\mathbf{c}$ is the global camera intrinsics.
The final energy function is optimized by the Gauss-Newton algorithm, which is further described in~\cite{engel_direct_2017, wang_stereo_2017}.

\subsubsection{Velocity Calculation}
After all stereo frames in the active window are optimized, the current velocity can be calculated from the poses of two latest stereo keyframes. A point is transformed from frame $I_i$ to frame $I_j$:
\begin{equation}
\displaystyle
\mathbf{T}_{ji} = \mathbf{T}_{j}^{-1}\mathbf{T}_i = 
\begin{bmatrix}
\mathbf{R}_{ji}   & t_{ji} \\
0       & 1 \\
\end{bmatrix}.
\end{equation}
\noindent
Rotational and translational velocity can be calculated from $\mathbf{R}_{ji}$ and $t_{ji}$, respectively. The rotational velocity $\omega$ can be defined as
\begin{equation}
\displaystyle
\omega = \atantwo \left( \mathbf{R}_{ji}(2,0), \sqrt{\mathbf{R}_{ji}(2,1)^2 + \mathbf{R}_{ji}(2,2)^2 } \right) / \Delta t,
\end{equation}
and the translational velocity $v$ can be defined as
\begin{equation}
\displaystyle
v = \sqrt{t_{ji}(0)^2 + t_{ji}(2)^2 } / \Delta t,
\end{equation}
where $\mathbf{R}_{ji}$ is 3-by-3 matrix, $v$ is a 3-by-1 vector.

\subsection{Cognitive Map}
The topological map can be optimized by solving a non-linear least squares problems~\cite{zeng_compactmapping_2017}. The normal equations are solved by a sparse solver with high performance, Ceres solver~\cite{agarwal2012ceres}. Nodes are introduced to denote poses of the robot. Links are utilized to model spatial constraints between nodes. The energy function of global optimization of topological map can be written as 
\begin{equation}
\begin{split}
\displaystyle
\min_{\mathbf{e}} &\quad \frac{1}{2}\sum_{i,j} \rho_i\left(\left\|f_i\left(e_{i},e_{j},e_{ij}\right)\right\|^2\right),
\end{split}
\end{equation}
where, $\mathbf{e}$ including all nodes $e_i$ and $e_j$ is optimized given links $e_{ij}$, $e_i = (x_i,y_i,\theta_i)$ and $e_j = (x_j,y_j,\theta_j)$ are the poses of the robot. $e_{ij} = (x_{ij},y_{ij},\theta_{ij})$ describes the constraint from $e_i$ to $e_j$.  $\rho_i\left(\left\|f_i\left(e_{i},e_{j},e_{ij}\right)\right\|^2\right)$ is a residual block, where $f_i(\cdot)$ is a cost function. $\rho_i$ is a loss function, namely Huber Loss, which largely weakens the influence of outliers during the process of the global optimization of the topological map. Cost function $f_i(\cdot)$ can be more specifically written as  
\begin{gather}
\begin{split}
f_i\left(e_{i},e_{j},e_{ij}\right) 
&= \begin{bmatrix} 
e_j - e_i - e_{ij} 
\end{bmatrix} 
= \begin{bmatrix} 
x_j - x_i - x_{ij} \\ 
y_j - y_i - y_{ij} \\ 
\theta_j - \theta_i - \theta_{ij} 
\end{bmatrix}\\
&= \begin{bmatrix}
x_j - x_i - d_{ij} \cdot \cos(\theta_i + \text{heading\_rad}) \\ 
y_j - y_i - d_{ij} \cdot \sin(\theta_i + \text{heading\_rad}) \\ 
\theta_j - \theta_i - \text{facing\_rad}
\end{bmatrix}, \\
\text{s.t.} &\quad -\pi \le \theta_i < \pi ,\\
&\quad -\pi \le \theta_j < \pi,
\end{split}
\end{gather}
where $e_{ij}$ describes the spatial constraint between $e_i$ and $e_j$, heading radians $\textit{heading\_rad}$, and facing radians $\textit{facing\_rad}$, $d_{ij}$ is the distance between $e_i$ and $e_j$. As relative angle radians do exist when visual template matching, heading radians are different from facing radians~\cite{ball_openratslam:_2013}. Values of $\theta_i$ and $\theta_j$ are limited to $[-\pi,\pi)$. 

\subsection{Implementation of Neurobiological Inspired SLAM system}

\begin{figure}[!ht]
\centering
\includegraphics[width=3.5in]{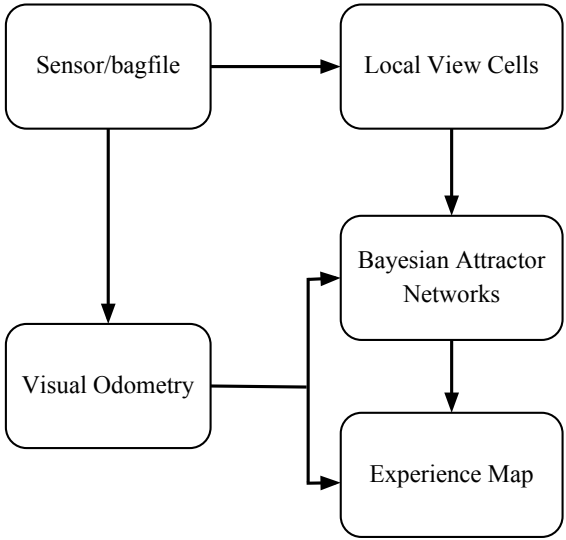}
\caption{The software architecture of the neurobiological inspired SLAM system. The sensor/bagfile node only provides image sequences. Visual odometry provides velocity estimation from pure stereo visual inputs. The local view cell node determines whether the current view is novel or not. Bayesian attractor network node performs path integration and makes decisions of loop closures. The experience map node generates the topological map.}
\label{fig_node_struct}
\end{figure}

Our neurobiological inspired SLAM system is modularly implemented in C++ language by Robot Operating System (ROS) with visualization. The software architecture are organized into five nodes shown in Fig.~\ref{fig_node_struct}, following the publicly available open-source OpenRatSLAM system~\cite{ball_openratslam:_2013}.

Python scripts are written to create a rosbag file with compressed stereo images from the KITTI odometry benchmark dataset sequences images and timestamps. The sensor/bagfile node real-time publishes compressed stereo images message to the local view cells node and the visual odometry node at 10 frames/s as they were recorded.

The visual odometry node estimates rotation information and linear speed by comparing consecutive keyframes with direct sparse visual methods from a moving stereo camera.

The local view cell node compares the current view from the left camera with visual temples to determine whether the robot enters a familiar location or not. When the robot revisits a familiar location, the corresponding local view cell can be coactivated with the HD cells and grid cells firing patterns through learned weights.

The Bayesian attractor network node receives odometry and view templates as inputs in the way of ROS message. The Bayesian attractor network comprises the HD cell network and the grid cell network. The ring attractor manifold of the HD cell network is described by an integrator cell and a calibration cell in one-dimensional Gaussian distribution. The torus attractor manifold of the grid cell network is also described by an integrator cell and a calibration cell in two-dimensional Gaussian distribution.
The vestibular cues drive the neural activity moving along the ring attractor manifold in the HD cell network and the torus attractor manifold in the grid cell network to represent the current pose of the robot. The decision about whether creating a new experience or closing a loop is made and is sent to the experience map node.

The experience map node builds a coherent semi-metric topological map. The vertexes define the pose of the robot. The edges describe the spatial constrains between two vertexes. Every vertex is described by a local view cell, head direction neural bump, and grid pattern. The location of the robot is mapped onto a torus, which means that an infinite area can be represented by grid cells. When the distance between the current location on the ring and torus manifold and the previous one is greater than a threshold, a ROS message is received from the Bayesian attractor network to create a new vertex and a new edge connecting the current vertex and the previous vertex. When loop closure occurs, a new edge would be created to connect an existing vertex. Then, the global optimization of the topological map is performed by Ceres Solver. A map is finally presented to represent the physical environment. 

The python scripts are written to visualize the live state of our neurobiological inspired SLAM system. The neural activity of HD cells and grid cells, the image view and the local view templates, and the experience map are all shown in our running system.

\section{Results}
\label{results}

We evaluated our neurobiologically inspired stereo visual SLAM system on the KITTI odometry benchmark dataset~\cite{Geiger2012CVPR}, and compared to the previous brain-inspired OpenRatSLAM with rough velocity estimation and neurobiologically inspired monocular visual SLAM system. 
The KITTI odometry benchmark dataset is recorded from a car with relatively high speed in urban and highway environments. The stereo camera has a $\sim$54cm baseline and works at 10Hz with a resolution of $1241 \times 376$ pixels.
We run our neurobiologically inspired stereo visual SLAM system on a six Intel Core i7-4930K personal computer with 64GB RAM. Video S1 in Supplementary Materials shows the mapping process of neurobiologically inspired stereo visual SLAM system.

\subsection{Neural Representation}

\begin{figure}[!ht]
\centering
\includegraphics[width=3.5in]{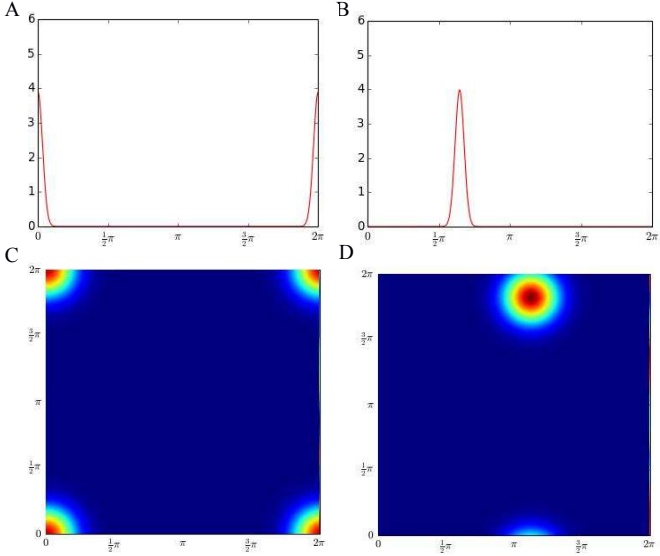}
\caption{Neural activities of HD cells and grid cells in the neurobiological inspired SLAM system. (A) and (C) show the initial state of the neural activities for HD cells and grid cells, respectively. (B) and (D) show one of the intermediate states of the neural activities for HD cells and grid cells, respectively. The activity bump of HD cells is centered at 2.02, about 115.64 degrees in (B). The activity bump of grid cells is centered at (3.56, 5.68) in (D).}
\label{fig_KITTI_neuralRep}
\end{figure}

The ring attractor manifold of HD cells is represented by a one-dimensional Gaussian distribution. The torus attractor of grid cells is represented by a two-dimensional Gaussian distribution with periodic boundary conditions, shown in Fig.~\ref{fig_KITTI_neuralRep}. At the beginning of the experiment, the robot is at the origin, and the neural activity of HD cells and grid cells is shown in Fig.~\ref{fig_KITTI_neuralRep}A and C, respectively. The Fig.~\ref{fig_KITTI_neuralRep}B and D show one of the intermediate states of the networks during the running process. In Fig.~\ref{fig_KITTI_neuralRep}B, the phase of the HD cell network is at 2.02, which suggests that the robot is currently heading at 115.64 degrees relate to the original head direction. The location of the robot represented by grid cells is centered at (3.56, 5.68). Due to the periodic boundary conditions on the torus manifold, the location of the robot is ambiguous.

\subsection{Cognitive Map}
\begin{figure}[!ht]
\centering
\includegraphics[width=3.3in]{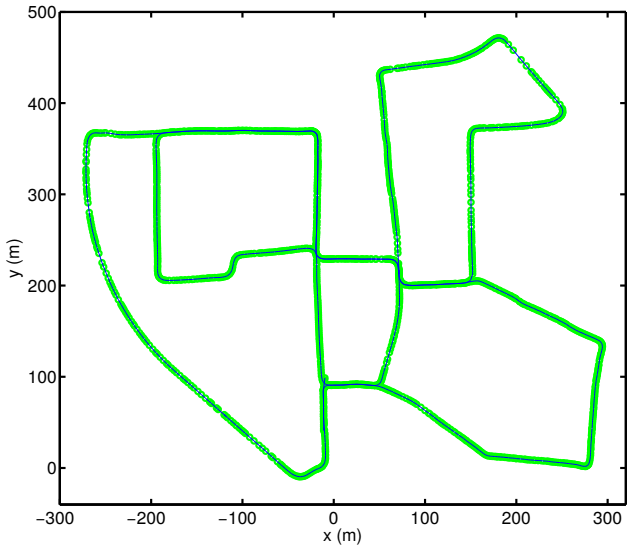}
\caption{The semi-metric topological map of the KITTI odometry benchmark dataset sequence 00 built by the neurobiologically inspired stereo visual SLAM system. The topological vertices is presented by the small green  circles. The blue thin line describes links between connected vertices.}
\label{fig_KITTI_cognitivemap}
\end{figure}

The cognitive map of the KITTI odometry benchmark dataset sequence 00 created by the neurobiologically inspired stereo visual SLAM is shown in Fig~\ref{fig_KITTI_cognitivemap}. The green line consists of a set of green vertices, which represent the robot position in the explored environment. The link connects two related vertices is presented by the fine blue line. Although there are small amount of loop closures and intersections in the KITTI odometry benchmark dataset sequence 00, the topological map captures the overall layout of the road network. 

\subsection{Local View Cells Activity}

\begin{figure}[!ht]
\centering
\includegraphics[width=3.5in]{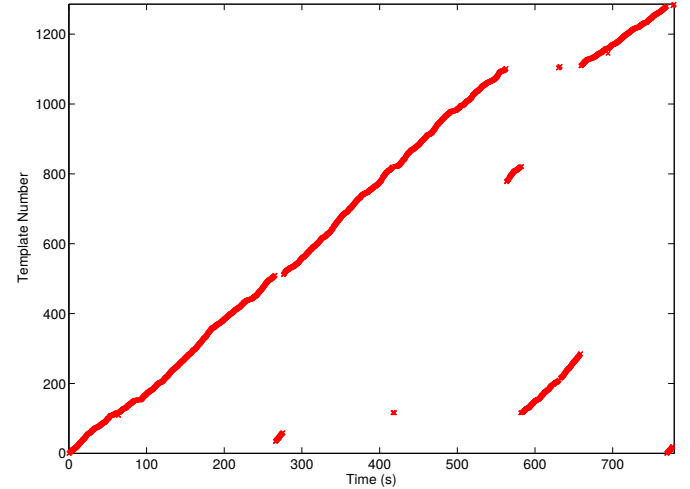}
\caption{Graph of the visual template over the time of exploration. }
\label{fig_KITTI_localviewactivity}
\end{figure}

There are 1286 visual templates learned during the mapping process of the KITTI odometry benchmark dataset sequence 00, as shown in Fig.~\ref{fig_KITTI_localviewactivity}. 
New view templates are continuously created as the upper bounding line. The familiar view templates are recognized as the short segments under the bounding line, which correspond to loop closures.


\subsection{Firing Rate Maps}

\begin{figure}[!ht]
\centering
\includegraphics[width=3in]{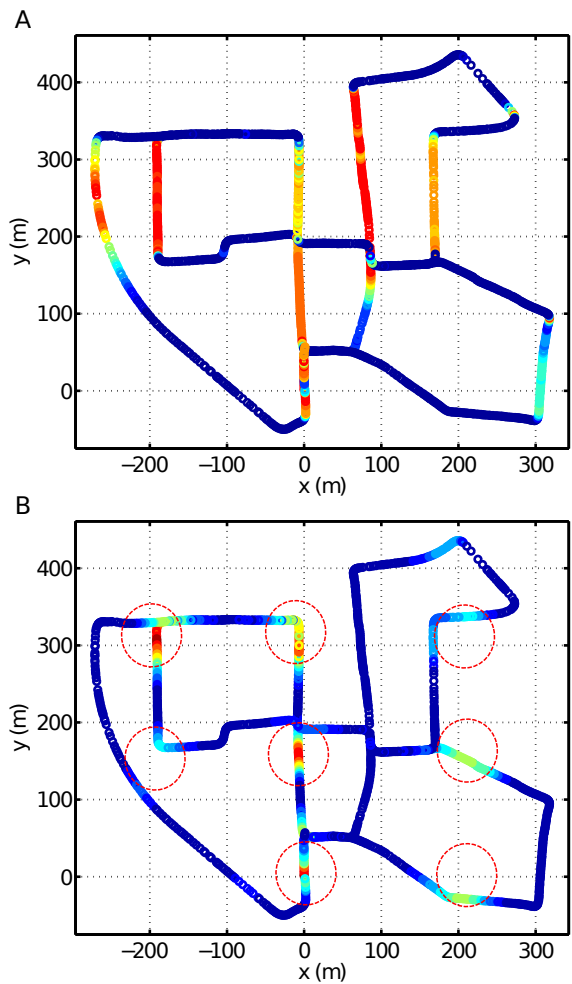}
\caption{Firing rate maps of HD cells and grid cells. For each panel, the same jet colormap is used to present the firing rate by color, ranging from blue (zero firing rate) to red (high firing rate). (A) The firing rate map of the total activity of the HD unit at 1 and 18 represents two opposite preferences in head direction, phase $0$ and $\pi$. (B) The firing rate map of the grid unit at (1,1) corresponds to the phase (0,0). The grid firing fields are labeled by dotted red circles.}
\label{fig_KITTI_firingratemaps}
\end{figure}

\begin{figure*}[!ht]
\centering
\includegraphics[width=7in]{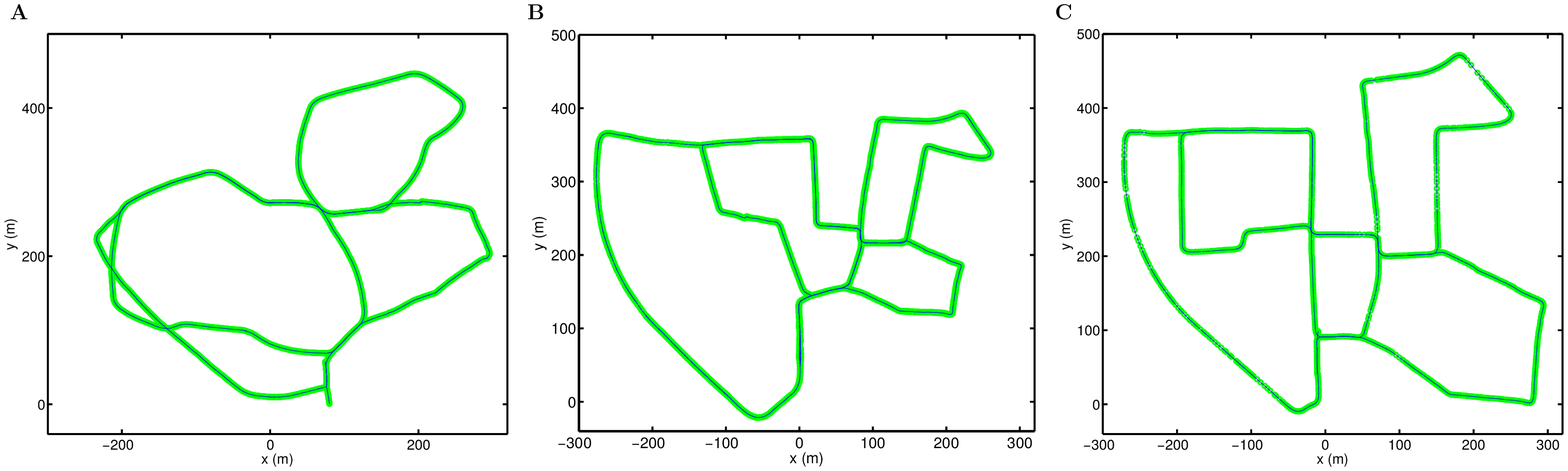}
\caption{Comparison of cognitive maps formed by our neurobiologically inspired SLAM system with the roughly estimated visual odometry (A), monocular visual odometry (B), and stereo visual odometry (C).}
\label{fig_KITTI_cognitivemap_comparison}
\end{figure*}

Firing rate maps not only show how HD cells and grid cells encode the explored environment, but also reflect the mapping performance of a cognitive model of spatial navigation. Fig.~\ref{fig_KITTI_firingratemaps}A shows the firing rate map of HD unit with label 1 and 18 in the ring attractor manifold. It encodes two opposite preference in head direction, corresponding to $0$ and $\pi$. The map in Fig.~\ref{fig_KITTI_firingratemaps}A shows a strong firing rate only when the robot moves north to south or south to north. Due to a little error accumulation of path integration, the firing rate intensity is slightly different between parallel paths. When the robot turns its head slowly, the firing rate intensity would gradually increase from zero, and then gradually decrease to zero. 

Since all grid units on the torus attractor manifold are limited by the periodic boundary conditions, each grid unit fires at multiple locations in the environment, shown in Fig.~\ref{fig_KITTI_firingratemaps}B. The grid unit $(1,1)$ encodes the spatial phase $(0,0)$ in the torus neural manifold. When the robot gets closer to the firing field center, the firing rate always increases. On the contrary, the firing rate always decreases.
The grid unit has about eight firing fields in the explored environment, and the center of these firing fields can form a square grid pattern, as we labeled these firing fields with eight dashed red circles.

As the cognitive map generated by the SLAM system with stereo cameras is barely distorted, HD cells only fire at a specific direction and grid cells express a square grid pattern in the large environment.

\subsection{Evaluation}

We qualitatively compared neurobiologically inspired SLAM system with stereo visual odometry against the SLAM system with monocular visual odometry and the roughly estimated visual odometry used in OpenRatSLAM, whose mapping processes are shown by video S1, S2, and S3 in Supplementary Materials, respectively.
The cognitive maps are shown in Fig~\ref{fig_KITTI_cognitivemap_comparison}. 
The rough topological map is shown in Fig~\ref{fig_KITTI_cognitivemap_comparison}A. Since visual odometry just makes a rough estimation, the layout of the road network cannot even be recognized. Nevertheless, the four loop closures and intersections are all conserved in the topological map. 
The SLAM system with a monocular camera significantly suffers the scale drift problem. The cognitive map generated by the SLAM system with a monocular camera is geometrically distorted, shown in Fig~\ref{fig_KITTI_cognitivemap_comparison}B. Fortunately, the layout of the road network, loop closures, intersections, corners, and curves in the semi-metric map can be clearly seen using naked eyes. 
However, for the cognitive map built by the SLAM system with stereo visual odometry, shown in Fig~\ref{fig_KITTI_cognitivemap_comparison}C, it cannot easily tell the difference between our built semi-metric map and ground truth, without quantitative comparison. The SLAM system with stereo cameras apparently shows better performance than another two algorithms.

\begin{figure}[!ht]
\centering
\includegraphics[width=3in]{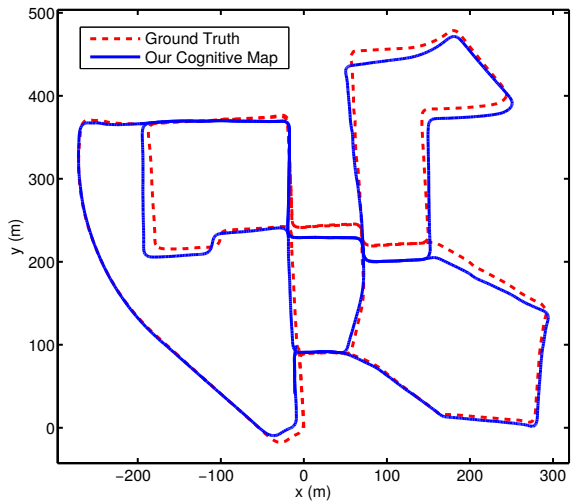}
\caption{Our cognitive map (Blue) and ground truth (Red) of sequence 00 from the KITTI odometry benchmark dataset.}
\label{fig_KITTI_cognitivemap_groundtruth}
\end{figure}

Then, we quantitatively compared our cognitive map generated by the neurobiologically inspired visual SLAM system with stereo camera against the ground truth, shown in Fig.~\ref{fig_KITTI_cognitivemap_groundtruth}. Considering the problem of time sequence alignment, we calculated root-mean-square error (RMSE) between our cognitive map and the ground truth. 
The MATLAB script was written to provide statistical analysis for the comparison. The mean error is 4.82 m, the median of the error is 4.50 m, and the RMSE error is 5.87 m. The minimum error and the maximum error are 0.03 m and 15.04 m, respectively. Consider dataset recorded by a car, whose length is more than 4.7 m, the quantitative comparison between our cognitive map and the ground truth is quite acceptable.

\subsection{More Results on KITTI}
\begin{figure}[!ht]
\centering
\includegraphics[width=3.6in]{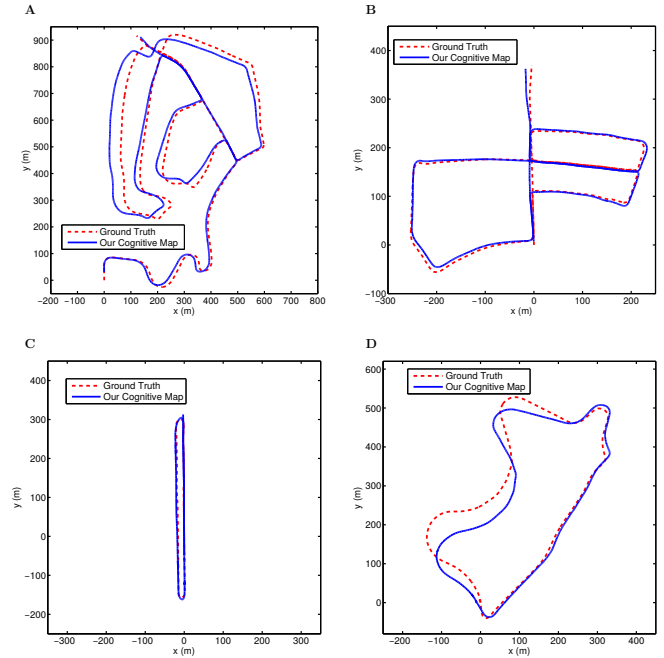}
\caption{Our cognitive maps in sequences 02, 05, 06 and 09 from the KITTI odometry benchmark dataset. (A) Sequence 02. (B) Sequence 05. (C) Sequence 06. (D) Sequence 09.}
\label{fig_morekitti}
\end{figure}

To further test our SLAM system, we demonstrated more sequences from the KITTI odometry benchmark datasets on the neurobiologically inspired stereo visual SLAM system. 
Qualitative comparisons of our topological maps and the ground truth are shown in Fig.~\ref{fig_morekitti}. Although there exist small errors between our topological maps and the ground truth, curves, corners, intersections, loop closures, and layout of maps are similar to the ground truth, which can be clearly seen from our qualitative comparison.

\section{Discussion}
\label{discussion}
We proposed a neurobiologically inspired stereo visual SLAM system based on a direct sparse method, which only uses visual information from a moving stereo camera to build cognitive maps in the large-scale environment. We demonstrated our SLAM system on the KITTI odometry benchmark datasets, which can successfully build accurate coherent topological semi-metric maps of challenging urban and highway environments recorded by a car with relatively high speed. The stereo visual SLAM system greatly outperforms the SLAM system with monocular visual odometry and the roughly estimated visual odometry used in OpenRatSLAM both in terms of tracking accuracy and robustness. Also, the firing rate maps provide evidence for the accuracy of the cognitive maps. 

The cognitive robot navigation system has long been known to validate biological models and test hypotheses. Here, we fuse the traditional visual algorithm to neurobiological inspired robot navigation systems, and further boost the performance of the robot system. Our neurobiologically inspired SLAM system can real-time run on the KITTI odometry benchmark dataset with high accuracy and robustness like traditional visual SLAM system (e.g. ORB-SLAM~\cite{mur-artal_orb-slam:_2015}).

Accuracy of the cognitive map generated by our neurobiological inspired SLAM system is a little lower than traditional algorithms, especially stereo DSO, for the following reasons. 
First, since we calculate current velocities from the latest two consecutive keyframes, the following frames cannot correct the previous velocity information.
Second, we only consider rotational and translational velocity relative to the ground, not including all velocity information like stereo DSO.
Third, considering that network attractor dynamic is applied to determine whether the robot revisits a familiar place, the process of network dynamic delays the loop closures.  

Also, there are some advantages in our neurobiological inspired SLAM system. 
First, we just estimate our velocity from two latest consecutive stereo images in the active window, and whereas, in the stereo DSO, all active points and frames are needed to be conserved for the following mapping and localization.
Second, in the experience map node, only vertices and links are needed to build the topological semi-metric map. Stereo DSO is required to store all 3D points in the robot mapping process. 
Third, since our mapping results is a topological map, it requires less memory and has high computational efficiency. It is more suitable for the large-scale environment.
Finally, the modular implementation of our SLAM system with ROS is beneficial for further developments and applications.

Some limitations also exist in our system. First, we sacrifice the metric accuracy for the cognitive map compared with traditional algorithms. Second, loop closures and intersections in environments of KITTI odometry benchmark dataset sequences are too few in number. Third, our SLAM system is not demonstrated in the real physical environment using a robot navigation system. 

In the future, we would deploy our SLAM system to the robot platform and demonstrate our SLAM system on the real physical environment.


\section{Conclusion}
\label{conclusion}

In brief, a neurobiologically inspired stereo visual SLAM system based on direct sparse method is proposed to build a coherent semi-metric map of the large-scale environment. Head direction cells and grid cells have the same mechanism to represent the head directions and positions of the robot in the physical environment. Traditional visual algorithm is fused into our neurobiologically inspired SLAM system to greatly improve the accuracy of the topological map. Our algorithm in this paper further facilitates brain-inspired SLAM to broader practical applications jointed with traditional methods.


\ifCLASSOPTIONcaptionsoff
  \newpage
\fi

\bibliographystyle{IEEEtran}
\bibliography{IEEEabrv,dsm}





\vfill



\end{document}